\documentclass[11pt]{article}
\usepackage{acl-hlt2011}
\usepackage{times}
\usepackage{latexsym}
\usepackage{amsmath}
\usepackage{multirow}
\usepackage{url}
\usepackage{graphics}
 \usepackage{graphicx}

\newenvironment{myitemize}{
\vspace{-0.5\baselineskip}
\begin{itemize}
\setlength{\topsep}{0pt}
\setlength{\itemsep}{0pt}
\setlength{\parskip}{0pt}
\setlength{\parsep}{0pt}
\setlength{\partopsep}{0pt}
}{
\end{itemize}
\vspace{-0.2\baselineskip}}
 \title{Colourful Language: Measuring Word--Colour Associations}

 \author{Saif Mohammad \\
 	Institute for Information Technology\\
   National Research Council Canada\\
     Ottawa, Ontario, Canada, K1A 0R6\\
     {\tt saif.mohammad@nrc-cnrc.gc.ca}
 }
\date{}

\begin{document}
\maketitle
\begin{abstract}
% Colour is a key component in the successful dissemination of information.  
Since many real-world concepts are associated with
colour, for example {\it danger} with red, linguistic information is
often complimented with the use of appropriate colours in information
visualization and product marketing.  Yet, there is no comprehensive
resource that captures concept--colour associations.  We present a
method to create a large word--colour association lexicon by
crowdsourcing. 
% We use a word-choice question to do sense-level annotations and to ensure data quality.  
We focus especially on
abstract concepts and emotions to show that even though they cannot be
physically visualized, they too tend to have strong colour
associations.  Finally, we show how word--colour associations manifest
themselves in language, and quantify usefulness of co-occurrence and
polarity cues in automatically detecting colour associations.\footnote{This paper is an extended,
non-archival, version of the short paper---Mohammad \shortcite{Mohammad11a}. It provides additional details on
the analysis of crowdsourced data, and experiments on the manifestations of word--colour associations in WordNet and in text.
It also proposes a polarity-based automatic method.}

% Thus, using the right colours can not only improve
% semantic coherence, but also inspire the desired psychological
% response.
% We show that more than 30\% of the terms have a strong colour association.
% There is no correlation between imageability and colour association,
% whereas emotions have  strong colour associations. 
% We also found that preference for colours has a remarkable correlation with
% the order in which colour terms appeared in language.
\end{abstract}

\section{Introduction}
Colour is a vital component in the successful delivery of information, whether it is
in marketing a commercial product \cite{SableA10}, designing webpages \cite{Meier88,Pribadi90}, or visualizing information \cite{Christ75,CardMS99}.
% This is because
Since real-world concepts have associations with certain colour categories
(for example, {\it danger} with red, and {\it softness} with pink),
complimenting linguistic and non-linguistic information with appropriate colours has a number of benefits, including:
(1) strengthening the message (improving semantic coherence), 
(2) easing cognitive load on the receiver,
(3) conveying the message quickly, and
(4) evoking the desired emotional response.
Consider, for example, the use of red in stop signs. % In North America, stop signs % have ``STOP" written on them, but they 
% are coloured red. Red has an association with danger, and so strengthens the information
% intended to be delivered by the sign. 
Drivers are able to recognize the sign faster, and it evokes a subliminal emotion pertaining to danger, which is entirely appropriate in the context.
% for a stop sign whose purpose is to avoid accidents.
The use of red to show areas of high crime rate in a visualization is another example of good use of colour to draw emotional response.
On the other hand, improper use of colour can be more detrimental to understanding than using no colour \cite{Marcus82,Meier88}.

% In a visualization of the effect of climate change on glaciers, snow sheets are best shown in white.

% Using the right colours is crucial in marketing a commercial product, in web design,
% For example, pinks are used with products targeting young girls. colours are also important 
% and in information visualization. 
% Visualizations are a graphical representation intended to give insights into data. The data pertains to different concepts, and 
% Using the right colours to represent different concepts will make it easier for the consumer to follow the information.
% For example, 
% Appropriate colours can also be used to convey the appropriate emotional effect a visualization or website is intended to evoke. 
% in a visualization of electoral results of USA, blue is better suited to represent Democrats and red for Republicans.

Most languages have expressions involving colour, and many of these express sentiment. Examples in English include:
{\it green with envy}, {\it blue blood} (an aristocrat), {\it greener pastures} (better avenues),
{\it yellow-bellied} (cowardly), {\it red carpet} (special treatment), and {\it looking through rose-tinted glasses} (being optimistic).
% Such expressions are not a fixed list, and 
Further, new expressions are continually coined, for example,
{\it grey with uncertainty} from Bianca Marsden's poem {\it Confusion}.\footnote{http://www.biancaday.com/confusion.html}
Thus, knowledge of concept--colour associations may also be useful for automatic natural language systems
such as textual entailment, paraphrasing, machine translation, and sentiment analysis.

 A word has strong association with a colour when the colour is a salient feature of the concept
 the word refers to, or because the word is related to a such a concept.
Many concept--colour associations, such as {\it swan} with white and {\it vegetables} with green, involve
physical entities.
% may be easy to recall,
% but there are innumerable concepts in the world. % and many of them may have associations with colours.
% The associated concepts are not restricted to physical entities such as snow and plants,
% but can also refer to 
However, even abstract notions %, such as {\it honesty} and {\it danger}, 
and emotions
% such as {\it joy} and {\it anger}, 
may have colour associations ({\it honesty}--white, {\it danger}--red, {\it joy}--yellow, {\it anger}--red).
% In fact, there are far more concepts in the world than there are colour categories, which
% means that the same colour may be associated with many concepts. For example, red
% is associated with danger, passion, and republicans, to name a few.
Further, many associations are culture-specific \cite{Gage99,Chen05}. For example, {\it prosperity} is associated with
red in much of Asia.
% Not every concept is associated with a colour, and different concepts may be associated
% to colours to a different degree. 

Unfortunately, there exists no lexicon with any significant coverage that captures concept--colour associations,
and a number of questions remain unanswered, such as, the  extent to which humans agree on these associations,
and whether physical concepts are more likely to have a colour association than abstract ones.
We expect that the word--colour associations manifest themselves as co-occurrences
in text and speech, but there have been no studies to show the extent to which words co-occur more with associated colours
than with other colours.

In this paper, we describe how we created a large word--colour association lexicon by crowdsourcing with effective quality control measures (Section 3).
We used a word-choice question to guide the annotators toward the desired senses of the target words, and also
to determine if the annotators know the meanings of the words. % she is giving colour associations for.
% We will make the lexicon publicly available.

We conducted several experiments to measure the consensus in word--colour associations, and how these associations manifest themselves in language.
Specifically, we show that:
% We conducted experiments to determine if physical entities that can be visualized tend to have strong colour associations more often
% than abstract concepts.
% We determined the extent to which concept-colour associations manifest themselves as co-occurrence in text 
% (does a concept co-occur most with the colour it is associated with, than any other colour), and how often the associations
% manifest as closeness in a semantic network, such as WordNet (is the target concept synset closest to a synset of the colour
% it is associated with, than any that of any other colour).
% Finally, we determined which colours are most strongly associated with different emotion words.
% We found that:

% We propose a method to determine the non-abstractness of thesaurus categories using 
 \begin{myitemize}
 \item More than 30\% of terms % and more than blah\% of thesaurus categories 
 have a strong colour association (Sections 4). 
% \item There is a fair amount of agreement among humans for word--colour associations (Section 4).
% \item Frequencies of colour choice in associations follows the same order in which colour terms ﬁrst appeared in language.
\item  % We use the groupings of words in a thesaurus to show that even 
% Even thesaurus groupings of related words have strong colour associations (Section 5).
About 33\% of thesaurus categories have strong colour associations (Section 5).
%  \item There is a definite order of preference for colours in associations. 
% \item If the colours are ranked in decreasing order of the number of times they were picked to be associated with a word,
%  we arrive at the same order in which colour terms occur in most languages \cite{BerlinK69} (Section 4).
% \item Even though it may seem that a larger percentage of concrete and imageable terms may have strong colour associations,
% \item There is very little, if any, correlation between the imageability of terms and their tendency to have an association with colour. In other words, 
\item Abstract terms have colour associations almost as often as physical entities do (Section 6).
% \item Strongly associated concept--colour pairs do co-occur with each other more often than chance, but
% even for the strongest of concept--colour associations, these co-occurrence cues are often not sufficient in determining which colour is most associated with a concept.
% \item The co-occurrence cues are almost equally strong for both abstract and concrete physical entities. (The cues are not markedly stronger for physical entities.)
\item There is a strong association of emotions and polarities with colours (Section 7).
% \item Some colours are predominantly positive whereas others are predominantly negative (Section 7).
\item Word-colour association manifests itself as closeness in WordNet (to a smaller extent), and as high co-occurrence in text (to a greater extent) (Section 8).
% \item Frequency-based ranking of colour terms in the
% {\it British National Corpus (BNC), Google N-gram Corpus (GNC)}, and {\it Google Books Corpus (GBC)}  
% have high correlations with the Berlin and Kay order (Section~8).
\end{myitemize}
\noindent Finally, we present an automatic method to determine word--colour association that relies on co-occurrence and polarity cues, but
no labeled information of word--colour associations.
It obtains an accuracy of more than 60\%. Comparatively, the random choice and most-frequent class supervised
baselines obtain only 9.1\% and 33.3\%, respectively.
% We expect semi-supervised approaches that use seed word--colour associations would obtain even better results.
Such approaches can be used to  % further improve the coverage of the manually created lexicon, and 
for creating similar lexicons in other languages.
% Frequencies of colour choice in associations follows the same order in which colour terms ﬁrst appeared in language.

% -- Imagability\\
% -- Agreement\\
% - - - Art\\
% web design

\section{Related Work}
 The relation between language and cognition has received considerable attention
 over the years, mainly on answering whether language impacts thought,
 and if so, to what extent. Experiments with colour categories have been used both to show
 that language has an effect on thought \cite{BrownL54,Ratner89}
 and that it does not \cite{Bornstein85}. However, that line of work does not
explicitly deal with word--colour associations. 
In fact, we did not find any other academic work that gathered large word--colour associations. There is, however,
a commercial endeavor---Cymbolism\footnote{http://www.cymbolism.com/about}.

%  Luscher \shortcite{Luscher69} argued that there is a correlation between
%  personal preference for certain colours and one's internal psychological state.

Child et al.\@ \shortcite{Child68}, Ou et al.\@ \shortcite{Ou11}, and others show that people of different ages and genders
have different colour preferences. (See also the online study by Joe Hallock\footnote{http://www.joehallock.com/edu/COM498/preferences.html}.)
In this work, we are interested in identifying words that have a strong association with a colour
due to their meaning; associations that are not affected by age and gender preferences.

There is substantial work on inferring the emotions evoked by colour \cite{Luscher69,Xin04,Kaya04}.
%In Section 7, we determine the association of {\it emotional words} with colours.
Strapparava and Ozbal \shortcite{StrapparavaO10} compute corpus-based semantic similarity between emotions and colours.
 We combine the word--colour and word--emotion association lexicons to determine the correlation between 
 emotion-associated words and colours.
% We investigate the correlation of emotions with colours in our work.
% proposed a test wherein a person is asked to
% rank coloured cards in order of preference. He argued that the order of these colours provides
% insight into the psychological state of the person. 
% Since then there have been both proponents () and detractors () of the theory.
% We show that some thesaurus categories and emotions have strong colour associations,
% however, we do not argue that preference for certain colours is indicative of associated psycological states.

Berlin and Kay \shortcite{BerlinK69}, and later Kay and Maffi \shortcite{KayM99}, showed that often colour
terms appeared in languages in certain groups. If a language has only two colour terms, then they
are white and black. 
If a language has three colour terms, then they are white, black, and red.
If a language has four colour terms, then they are white, black, red, and green, and so on
up to eleven colours. From these groupings, the colours can be ranked as follows:
% Below is the most common order (earlier to later):
 \begin{quote}
1. white, 2. black, 3. red, 4. green, 5. yellow, 6. blue, 7. brown, 8. pink, 9. purple, 10. orange, 11. grey \hspace{2.7cm} (1)
\end{quote}
\noindent We will refer to the above ranking as the {\it Berlin and Kay (B\&K) order}.
There are  hundreds of different words for colours.\footnote{See http://en.wikipedia.org/wiki/List\_of\_colors}
 To make our task feasible, we needed to choose
 a relatively small list of basic colours. We chose to
 use the eleven basic colour words of Berlin and Kay (1969).

% We propose a method to determine the non-abstractness of thesaurus categories using 
The MRC Psycholinguistic Database \cite{Coltheart81} has, among other information, the {\it imageability
ratings} for 9240 words.\footnote{http://www.psy.uwa.edu.au/mrcdatabase/uwa\_mrc.htm}
The imageability rating is a score given by human judges that reflects
how easy it is to visualize the concept.
It is a scale from 100 (very hard to visualize) to 700 (very easy to visualize).
We use the ratings  in our experiments to determine whether there is a correlation between imageability and
strength of colour association.

\begin{table*}[]
\centering
% \resizebox{\textwidth}{!}{
 {\small
\begin{tabular}{l rrrr rrrr rrr}
\hline
      %   \multicolumn{11}{c}{\bf colour} \\
            & white &black &red &green &yellow &blue &brown &pink &purple &orange &grey\\
\hline
            overall     &11.9       &12.2       &11.7       &12.0       &11.0       &9.4        &9.6        &8.6        &4.2  &4.2 &4.6\\
            voted   &22.7    &18.4       &13.4       &12.1       &10.0       &6.4        &6.3        &5.3        &2.1        &1.5 &1.3\\
\hline
\end{tabular}
}
 \vspace*{-2mm}
\caption{ Percentage of terms marked as being associated with each colour.}
\label{tab:col votes}
% \vspace*{-3mm}
\end{table*}

\section{Crowdsourcing}

Amazon's Mechanical Turk (AMT) is an online crowdsourcing platform that is especially
well suited for tasks that can be done over the Internet through a computer
or a mobile device.\footnote{Mechanical Turk: www.mturk.com}
It is already being used to obtain human annotation
on various linguistic tasks \cite{SnowOJN08,CallisonBurch09}.
However, one must define the task carefully to obtain annotations of
high quality. Several checks must be placed to ensure that random and
erroneous annotations are discouraged, rejected, and re-annotated.

We used Mechanical Turk to obtain
word--colour association annotations on a large-scale. %\footnote{Mechanical Turk: www.mturk.com}
Each task is broken into small independently solvable units called {\it HITs
(Human Intelligence Tasks)} and uploaded on the Mechanical Turk
website.  
The people who provide responses to these HITs are called {\it Turkers}.  
The annotation provided by a Turker for a HIT is called an {\it assignment}.

We used the {\it Macquarie Thesaurus} \cite{Bernard86} as the source for terms to be annotated. % by people on Mechanical Turk.
Thesauri, such as the {\it Roget's} and {\it Macquarie}, group related words into categories.
The {\it Macquarie} has about a thousand categories, each having about a hundred or so related terms.
Each category has a {\it head word} that best represents the words in it.
The categories can be thought of as coarse senses or concepts \cite{Yarowsky92}.
If a word is ambiguous, then it is listed in more than one category.
Since a word may have different colour associations when used in different senses,
we obtained annotations at word-sense level.
% We were additionally interested in determining colour signatures for emotions (Section 7),
We chose to annotate words that had 
one to five senses in the {\it Macquarie Thesaurus} and occurred
frequently in the {\it Google N-gram Corpus}.
We annotated more than 10,000 of these word--sense pairs  % that Mohammad and Turney \shortcite{MohammadT10} used to create their emotion lexicon.
by creating HITs as described below. 

Each HIT has a set of questions, all of
which are to be answered by the same person.  
We requested annotations from five different Turkers for each HIT.
(A Turker cannot attempt multiple assignments for the same term.)
% Different HITS may be attempted by different Turkers, and Turkers may choose to attempt as many HITs as they wish.
% Since the {\it Roget Thesaurus} is available freely in the public domain, it allows us
% to distribute our emotion lexicon without the burden of restrictive licenses. 
% We chose only those words that occurred more than
% 120,000 times in the Google n-gram corpus.\footnote{The frequency threshold of 120,000 is arbitrary.
% We wanted to annotate the most frequent words first. The Google n-gram corpus is available through the Linguistic Data Consortium.}
%  We bias the annotator towards a particular sense of the target word by presenting a related word from the
%  target thesaurus category (target sense).
A complete HIT
% with directions 
is shown below:
% Below is an example questionnaire:\\

\rule{7.6cm}{0.4mm}
% \vspace*{-3mm}

{\small
% {\bf Title:} Colours associated with words.\\
% {\bf Keywords:} emotion, English, sentiment, word association, word meaning\\
% \indent {\bf Reward per HIT:} \$0.01
% \indent {\bf Directions:}
% \begin{myitemize}
% \item This survey will be used to better understand word--colour associations. Your input is appreciated.
% \item Please return/skip HIT if you do not know the meaning of the word.
% \item Attempt HIT only if you are a native speaker of English and you know the meaning of the prompt word. %, or very fluent in English.
% \item Certain ``check questions" will be used to make sure the annotation is responsible and reasonable. 
% \item Your responses are confidential. 
% \item Some words have more than one meaning, and the different meanings may be associated with different emotions.  For each HIT, Q1 will guide you to the intended meaning. Your input is appreciated.
% % You may encounter multiple HITs for the target word, and they will have different guiding questions.
   % \item Note that the order of options in Q13 changes with every HIT.
% \end{myitemize}

% {\bf Prompt word: {\it sleep}}

\vspace*{1mm}
\indent Q1. Which word is closest in meaning  % (most related) 
to {\it sleep}?

% \begin{minipage}[t]{5mm}
% \end{minipage}
% \begin{minipage}[t]{4cm}
% \begin{myitemize}
%\item {\it crocodile}
%\item {\it honesty}
%\item {\it nap}
%\item {\it olive}\\
%\end{myitemize}
% \end{minipage}

\vspace*{1mm}
 \begin{minipage}[t]{4mm}
 \end{minipage}
 \begin{minipage}[t]{1.7cm}
  \begin{myitemize}
 \item {\it car}
 \end{myitemize}
 \end{minipage}
 \begin{minipage}[t]{1.9cm}
  \begin{myitemize}
 \item {\it tree}
 \end{myitemize}
 \end{minipage}
 \begin{minipage}[t]{1.7cm}
  \begin{myitemize}
 \item {\it nap}
 \end{myitemize}
 \end{minipage}
 \begin{minipage}[t]{1.6cm}
 \vspace*{-0.7mm}
  \begin{myitemize}
 \item {\it olive}\\
 \end{myitemize}
 \end{minipage}

\vspace*{-3mm}
Q2. What colour is associated with {\it sleep}?

% \vspace*{-1mm}
% \begin{minipage}[t]{5mm}
% \end{minipage}
%\begin{minipage}[t]{2.5cm}
% \begin{myitemize}
% \item black
% \item blue
% \item brown
% \item green
% \item grey
% \item orange
%\end{myitemize}
%\end{minipage}
%\begin{minipage}[t]{2.5cm}
% \begin{myitemize}
% \item purple
% \item pink
% \item red
% \item white
% \item yellow\\
%\end{myitemize}
%\end{minipage}

 \vspace*{1mm}
  \begin{minipage}[t]{4mm}
  \end{minipage}
 \begin{minipage}[t]{1.7cm}
  \begin{myitemize}
  \item black
  \item blue
  \item brown
 \end{myitemize}
 \end{minipage}
 \begin{minipage}[t]{1.8cm}
  \begin{myitemize}
  \item green
  \item grey
  \item orange
 \end{myitemize}
 \end{minipage}
 \begin{minipage}[t]{1.7cm}
  \begin{myitemize}
  \item purple
  \item pink
  \item red
 \end{myitemize}
 \end{minipage}
 \begin{minipage}[t]{1.9cm}
  \begin{myitemize}
  \item white
  \item yellow\\
 \end{myitemize}
 \end{minipage}

\rule{7.6cm}{0.4mm}\\
}

\vspace*{1mm}
\noindent 
Q1 is a word-choice question generated automatically by taking a near-synonym
from the thesaurus and random distractors. 
The near-synonym also guides the annotator to the desired sense of the word.
Further, it encourages the annotator to think clearly about the target word's meaning; we believe this improves the quality of the annotations in Q2.
 If a word has multiple senses, that is, it is listed in more than one thesaurus category,
 then separate questionnaires are generated for each sense.
 Thus we obtain colour associations at a word-sense level.

If an annotator answers Q1 incorrectly,
then we discard information obtained from both Q1 and Q2. 
Thus, even though we do not have correct answers to Q2, likely incorrect annotations are filtered out.
% from malicious Turkers who deliberately enter incorrect information is 
About 10\% of the annotations were discarded because of an incorrect answer to Q1. 
Terms with less than three valid annotations were removed from further analysis.
Each of the remaining terms had, on average, 4.45 distinct annotations.

The colour options in Q2 were presented in random order.
Observe that we do not provide a ``not associated with any colour" option. This encourages
colour selection even if the annotator felt the association was weak. If there is no association
between a word and a colour, then we expect low agreement amongst the annotators.
% We requested annotations from five different people for each term.
% question makes the annotator think about the meaning of the target word, which we think helps in getting better annotations for Q2.
 The survey was approved by the ethics board at the authors' institution.
% at the National Research Council Canada.

% \section{Running on MTurk and post-processing}

\section{Word--Colour Association}

% For 74.4\% of the those instances, all five annotators agreed on whether a term is associated
% with a particular emotion or not. For 16.9\% of the instances four out of five people agreed with each other,
% and for 8.5\% of the instances there was a three--two split.
The information from multiple annotators % for a particular term 
was combined
by taking the majority vote, resulting in a lexicon of 8,813 entries.
% The lexicon has 8,813 entries. % (we will increase this to about 15,000 in the near future).
Each entry contains a unique word--synonym pair (from Q1), majority-voted colour, and
a confidence score---number of votes for the colour
/ number of total votes.
(For the analyses in the rest of the paper, ties were broken by picking one colour at random.)
A separate version of the lexicon that includes entries for all of the valid annotations by each of the annotators
is also available.\footnote{Please contact the author to obtain a copy of the lexicon.}

The first row, {\it overall}, in Table \ref{tab:col votes} shows the percentage of times different colours were
associated with the target term. % The first row shows raw percentages.
The second row, {\it voted},  shows percentages after taking a majority vote from multiple annotators.
% For each term, the colour that gets the majority votes (from the 3 to 5 annotators
% for that HIT) is chosen
% as the true colour associated with the target term, and the ``voted" row
% of Table \ref{tab:col votes}
% shows the percentage of terms associated with the different colours.
Observe that even though the colour options were presented in random order,
the order of the most frequently associated colours
is identical to the Berlin and Kay order
(Section 2:(1)).

Table \ref{tab:col agreement} shows how often the size of the majority class in colour associations
is one, two, three, four, and five.
Since the annotators were given eleven colour options to choose from, 
if we assume independence, then the chance that none of the five annotators agrees with each other
(majority class size of one) is $1 \times 10/11 \times 9/11 \times 8/11 \times 7/11 = 0.344$.
Thus, if there was no correlation among any of the terms and colours, then 34.4\% of the time
none of the annotators would have agreed. However, this happens only 15.1\% of the time.
A large number of terms have a majority class size $\geq$ 2 (84.9\%), and thus more than chance association with a colour.
One can argue that terms with a majority class size $\geq$ 3 (32\%)  have {\it strong} colour associations.
% For about 2.1\% of the words, all 5 annotators agree on the associated colours.

\begin{table}[t]
% \caption{colour agreement.}
\centering
% \resizebox{0.5\textwidth}{!}{
 {\small
\begin{tabular}{rrrrrrr}
\hline
        \multicolumn{7}{c}{\bf majority class size} \\
                    one     &two    &three   &four   &five &$\geq$ two &$\geq$ three\\
\hline
        15.1        &52.9        &22.4        &7.3     &2.1   &84.9 &32.0  \\
\hline
\end{tabular}
 }
 \vspace*{-2mm}
\caption{Percentage of terms in different majority classes.}
\label{tab:col agreement}
\end{table}

Below are some reasons why agreement values are much lower than those obtained for certain other tasks,
for example, part of speech tagging:
\begin{myitemize}
\item The annotators were not given a ``not associated with any colour" option.
Low agreement for certain instances is an indicator that these words have weak, if any, colour association.
\item Words are associated with colours to different degrees.
Some words may be associated with more than one colour in comparable degrees, and there might be higher disagreement for such instances.
% , and there are no clear classes corresponding to different levels of association.
% Since we ask people to place term-colour associations in specific bins (no association and association), more people disagree for term--colour pairs
% whose degree of association is closer to the boundary, than for term--colour pairs further away from the boundary.
% \item \cite{Holsti69}, \cite{BrennanP81}, \cite{PerreaultL89}, and others consider
% the $\kappa$ values (both Fleiss's and Cohen's) to be conservative, especially
% when one category is much more prevalent than the other.
% In our data, the ``not associated with emotion" category is much more prevalent than the ``associated with emotion" category,
% so these $\kappa$ values might be underestimates of the true agreement.
\item The target word is presented out of context. We expect higher agreement if
we provided words in particular contexts, but words can occur in innumerable contexts,
and annotating too many instances of the same word is costly.
%, and that will entail an explosion of instances
% to be annotated. 
% One can group contexts into those that correspond to different senses of the target word.
% By providing the word-choice question, we bias the Turker towards a particular sense
% of the target word, and aim to obtain the prior probability of the word sense's emotion association.
\end{myitemize}
\noindent Nonetheless, the term--colour association lexicon is useful for downstream applications because
any of the following strategies may be employed: (1) choosing colour associations from only those instances
with high agreement, (2) assuming low-agreement terms have no colour association, (3) determining colour association of
a category through information from many words, as described in the next section.
% , and use contextual information in addition to information obtained form the lexicon.

 \begin{figure*}[t]
  \begin{center}
  \includegraphics[width=1.4\columnwidth]{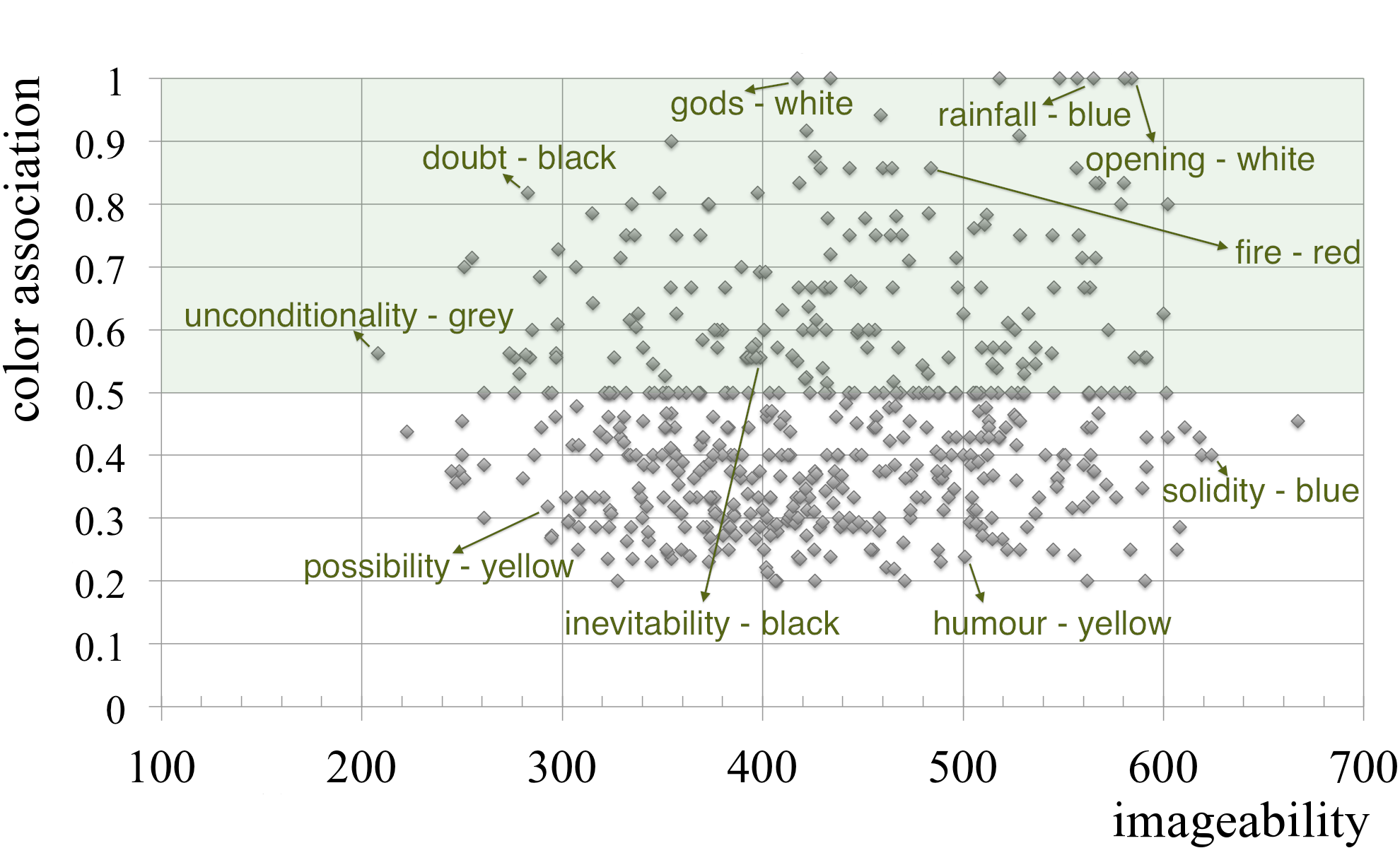}
  \end{center}
  % \caption{{\bf Polarity pie chart}---Proportions of polarities in quotations about gratitude. (Text from: Quote Garden)}
  % \caption{Scatter plot of imageability and association with colour of thesaurus categories.}
    \vspace*{-4mm}
  \caption{Scatter plot of thesaurus categories. The area of high colour association is shaded. Some points are labeled.}
  \label{fig:scatter}
  \vspace*{-3mm}
  \end{figure*}

\section{Category--Colour Association}

Words within a thesaurus category may not be strongly associated with any colour, or they may each be associated with many different colours.
We now describe experiments to determine whether there exist categories where the semantic coherence carries over to a strong common association with one colour. 

 We determine the strength of colour association of a category by first determining the colour $c$ most associated with the terms in it, 
 and then calculating the ratio of the number of times a word from the category is associated with $c$ to the number of words in the category associated with any colour. 
% The lowest possible strength of association is 1/11, and the highest is 1.
% Since our lexicon currently has entries for only about 10,500 words, 
% We calculated the strength of colour association of thesaurus categories using only those words appearing in in the word--colour lexicon. 
Only categories that had at least four words that also appear in the word--colour lexicon were considered; 535 of the 812  categories from  {\it Macquarie Thesaurus} met this condition. 

If a category has exactly four words that appear in the colour lexicon, and if all four words are associated with different colours, then the category has the lowest possible strength of colour association---0.25 (1/4). 
19 categories had a score of 0.25. No category had a score less than 0.25.
Any score above 0.25 shows more than random chance association with a colour. There were 516 such categories (96.5\%). 177 categories (33.1\%) had a score 0.5 or above, that is,
half or more of the words in these categories are associated with one colour. We consider these to be strong associations,
and a gold standard for automatic measures of association.
% The existence of so many (blah \%) such categories gives further credence to the assertion that certain concepts indeed have a strong colour association.
% Table \ref{tab:cat colours} shows the ten thesaurus categories having highest associations with colours.
% (The complete gold standard is submitted as supplementary material.)

% \begin{table}[]
% \centering
% \resizebox{0.5\textwidth}{!}{
% % {\small
%  \begin{tabular}{lll l lll}
% % \begin{tabular}{llll}
% \hline
% category  &colour  &score  &  &category  &colour  &score\\
% \cline{0-2} \cline{5-7}
% opening		&white	&1.00 	& &mariner     &blue   &1.00	\\
% rainfall	&blue	&1.00 	& &eloquence   &white  &1.00	\\
% marriage	&white	&1.00 	& &dive        &blue   &1.00	\\
% brown		&brown	&1.00 	& &gods        &white  &1.00	\\
% explosion	&red	&1.00 	& &cleanliness &white  &0.94	\\
% 
% \hline
% \end{tabular}
% }
% \caption{Categories with the highest colour associations.}
% \label{tab:cat colours}
% % \vspace*{-3mm}
% \end{table}

\begin{table*}[t]
\centering
% \resizebox{\textwidth}{!}{
 {\small
\begin{tabular}{l rrrr rrrr rrr}
\hline
%        &\multicolumn{11}{c}{\bf colour} \\
            &white &black &red &green &yellow &blue &brown &pink &purple &orange &grey\\
\hline

anger words           &2.1    &{\bf 30.7}   &{\bf 32.4}   &5.0  &5.0  &2.4    &6.6    &0.5    &2.3    &2.5    &9.9\\
anticipation words    &{\bf 16.2}   &7.5    &11.5   &{\bf 16.2}   &10.7   &9.5    &5.7    &5.9    &3.1    &4.9    &8.4\\
disgust  words    &2.0  &{\bf 33.7}   &{\bf 24.9}   &4.8    &5.5    &1.9    &9.7    &1.1    &1.8    &3.5    &10.5\\
fear words        &4.5    &{\bf 31.8}   &{\bf 25.0} &3.5    &6.9    &3.0  &6.1    &1.3    &2.3    &3.3    &11.8\\
joy words     &{\bf 21.8}   &2.2    &7.4    &{\bf 14.1}   &13.4   &11.3   &3.1    &11.1   &6.3    &5.8    &2.8\\
sadness words     &3.0  &{\bf 36.0} &{\bf 18.6}   &3.4    &5.4    &5.8    &7.1    &0.5    &1.4    &2.1    &16.1\\
surprise  words       &11.0 &13.4   &{\bf 21.0} &8.3    &{\bf 13.5}   &5.2    &3.4    &5.2    &4.1    &5.6    &8.8\\
trust words       &{\bf 22.0} &6.3    &8.4    &14.2   &8.3    &{\bf 14.4}   &5.9    &5.5    &4.9    &3.8    &5.8\\
\hline
\end{tabular}
 }
 \vspace*{-2mm}
\caption{Colour signature of emotive terms: percentage of terms associated with each colour.  For example, 32.4\% of the anger terms are associated with red.  The two most associated colours are shown in bold.}
\label{tab:col sig1}
 \vspace*{-1mm}
\end{table*}

% \vspace*{-14mm}
 \begin{table*}[t]
 \centering
 % \resizebox{\textwidth}{!}{
  {\small
 \begin{tabular}{l rrrr rrrr rrr}
 \hline
%          &\multicolumn{11}{c}{\bf colour} \\
             &white &black &red &green &yellow &blue &brown &pink &purple &orange &grey\\
 \hline
 
 negative        &2.9    		&{\bf 28.3} &{\bf 21.6}  	&4.7    		&6.9    	&4.1    	&{\bf 9.4}  &1.2    	&2.5    	&3.8    	&{\bf 14.1}\\
 positive        &{\bf 20.1}   	&3.9    	&8.0  			&{\bf 15.5}   	&{\bf 10.8} &{\bf 12.0} &4.8   		&{\bf 7.8}  &{\bf 5.7}  &{\bf 5.4}  &5.7\\
 \hline
 \end{tabular}
  }
 \vspace*{-2mm}
 \caption{Colour signature of positive and negative terms: percentage terms associated with each colour.  For example, 28.3\% of the negative terms are associated with black.  The highest values in each column are shown in bold.}
 \label{tab:col sig2}
 \vspace*{-1mm}
 \end{table*}

\section{Imageability and Colour Association}

It is natural for physical entities of a certain colour to be associated with that colour. However, abstract concepts such as {\it danger} and {\it excitability} are also associated with colours---red and orange, respectively.\\
{\mbox Figure \ref{fig:scatter}} displays an experiment to determine whether there is a correlation between imageability and association with colour.

We define imageability of a thesaurus category to be the average of
the  imageability ratings of words in it.  We calculated imageability
for the 535 categories described in the previous section using
only the words that appear in the colour lexicon.  Figure
\ref{fig:scatter} shows the scatter plot of these categories on
the imageability and strength of colour association axes. 
The colour association was calculated as described in the previous section.
% If there were a correlation between the two, that is 

If higher imageability correlated with
greater tendency to have a colour association, then we would 
see most of the points along the diagonal moving up from left to
right. Instead, we observe that the strongly associated categories (points in the shaded region)  are spread across the imageability axis,
implying that there is only weak, if any, correlation between
imageability and strength of association with colour.
Imageability and colour association have a Pearson's product
moment correlation of 0.116, and a Spearman rank order correlation of
0.102.

\section{The Colour of Emotion Words}

Emotions such as joy and anger are abstract concepts dealing with one's psychological state.
Mohammad and Turney \shortcite{MohammadT10} created a crowdsourced term--emotion association lexicon
consisting of associations of over 10,000 word-sense pairs with eight emotions---joy, sadness, anger, fear,
trust, disgust, surprise, and anticipation---argued to be the
basic and prototypical emotions \cite{Plutchik80}.
We combine their term--emotion association lexicon 
and our term--colour lexicon to
determine the colour signature of different emotions---the rows in Table \ref{tab:col sig1}.
 The top two most frequently associated colours with each of the eight emotions are shown in bold.
For example, the ``anger" row shows the percentage of anger terms associated with different colours. 

We see that all of the emotions have strong associations with certain colours. 
Observe that anger is associated most with red.
Other negative emotions---disgust, fear, sadness---go strongest with black.
% However, for these emotions black is much more dominant. 
% Grey comes third for all four of the negative emotions.
Among the positive emotions: anticipation is most frequently associated with white and green;
joy with white, green, and yellow; and trust with white, blue, and green.
% Surprise, which can be positive or negative, is associated with red, yellow, and black.
Thus, colour can add to the emotional potency of visualizations.

The Mohammad and Turney \shortcite{MohammadT10} lexicon also has associations with
positive and negative polarity. 
% These include terms that may not be associated with the eight basic emotions.
We combine these term--polarity associations with term--colour associations to
show the colour signature for positive and 
negative terms---the rows of Table \ref{tab:col sig2}.  
We observe that some colours tend to, more often than not, have strong positive associations (white, green, yellow, blue, pink, and orange),
whereas others have strong negative associations (black, red, brown, and grey).
% In Section 8.2, we will show how this information can be used as a cue in automatically determining
% the colour most associated with a term.

 \begin{table*}[t]
 \centering
 \resizebox{0.85\textwidth}{!}{
%  {\small
 \begin{tabular}{l rrrr rrrr rrr}
 \hline
 
%  	    &\multicolumn{11}{c}{\bf number of senses}\\
 colour 	& white 	&black 	&red 	&green 	&yellow 	&blue 	&brown 	&pink 	&purple 	&orange 	&grey\\ % 	& &\multicolumn{2}{c}{\bf correlation}\\	
% B\&K rank	&1		&2		&3		&4		&5			&6		&7		&8		&9			&10			&11\\ %		& &$r_{11}$				&$r_{6}$\\
% 	    &\multicolumn{11}{c}{frequency}  & &r (1--11)   &r (1--6)\\
%		\cline{1-12} \cline{14-15}
\hline
% noun 		&12		&7		&4		&8		&1		&7		&4		&3		&2		&5		&7\\ %		& &
% verb 		&1		&1		&0		&1		&1		&1		&2		&3		&2		&0		&2	\\ %	& &
% adjective  &12		&14		&3		&5		&6		&8		&2		&1		&3		&1		&4\\ %		&
% \hline
\# of senses		&25		&22		&7		&14		&8		&16		&8		&7		&7		&6		&13\\ %
 \hline
 \end{tabular}
  }
  \vspace*{-2mm}
 \caption{The number of senses of colour terms in WordNet.} % B\&K rank refers to the rank in the Berlin and Kay order.}
 \label{tab:colour senses}
  \vspace*{1mm}
 \end{table*}

 \begin{table*}[t]
 \centering
 \resizebox{0.97\textwidth}{!}{
%  {\small
 \begin{tabular}{lr rrrr rrrr rrr rr}
 \hline
 
&  			& white 	&black 	&red 	&green 	&yellow 	&blue 	&brown 	&pink 	&purple 	&orange 	&grey 	& &$\rho$\\	
\multicolumn{2}{r}{B\&K rank:}	&1		&2		&3		&4		&5			&6		&7		&8		&9			&10			&11		& &\\
%	\cline{1-2} \cline{3-13} \cline{15-16}
\hline
\multirow{2}{*}{\it BNC}  	&freq: 	&1480		&3460		&2070		&1990	&270	&1430	&1170	&450	&180	&360	&800	& &\rule{0pt}{14pt}			\\
&  rank:		&4			&1			&2			&3		&10		&5		&6		&8		&11		&9		&7		& &0.727			\\
\multirow{2}{*}{\it GNC} 	&freq: 	&205		&239		&138		&106	&80		&123	&63		&41		&16		&36		&18		& &\rule{0pt}{14pt} \\
&  	rank:	&2			&1			&3			&5		&6		&4		&7		&8		&11		&9		&10		& &0.884			\\
\multirow{2}{*}{\it GBC} &freq:  &233		&188		&130		&86		&44		&75		&72		&14		&11		&19		&22		& &\rule{0pt}{14pt} 		\\
&  rank:		&1			&2			&3			&4		&7		&5		&6		&9		&10		&11		&8		& &0.918			\\
 \hline
 \end{tabular}
  }
 \vspace*{-2mm}
 \caption{Frequency and ranking of colour terms per 1,000,000 words in the {\it British National Corpus (BNC), Google N-gram Corpus (GNC),} and {\it Google Books Corpus (GBC)}.  The last column lists the Spearman rank order correlation ($\rho$) of the rankings with the Berlin and Kay (B\&K) ranks.  }
 \label{tab:uni colours}
 \vspace*{-3mm}
% \vspace*{-3mm}
 \end{table*}

\section{Manifestation of Concept--Colour Association in WordNet and in Text}
% We now describe experiments to determine the extent to which the association of a 
% concept with a colour is observable in WordNet and in text corpora.

\subsection{Closeness in WordNet}
% WordNet % is a lexicographic resource that aims to 
% represents concepts as nodes in a network. The edge between two nodes represents
% a lexicographic relation, for example hypernymy or antonymy. Each node, called a {\it synset},
% has a set of near-synonymous words, and a brief definition called the {\it gloss}.
% A word with multiple senses is listed in multiple synsets. 
% Various measures, such as that proposed by Jiang and Conrath \shortcite{JiangC97} and Lin \shortcite{Lin97},
% can be used to estimate the semantic distance or closeness between two synsets.
Colour terms are listed in WordNet,
and interestingly, they are fairly ambiguous. Therefore, they can be found in many different synsets
(see Table \ref{tab:colour senses}). % lists the number of senses of each of the eleven basic colour terms.
A casual examination of WordNet reveals that some synsets (or concepts) are
close to their associated colour's synset. For example, {\it darkness} is a hypernym of black
and {\it inflammation} is one hop away from red.
It is plausible that if a concept is strongly associated with a certain colour, then 
such concept--colour pairs will be close to each other in a semantic
network such as WordNet.
% In order to determine the extent to which the above proposition is true, we calculated
If so, the semantic closeness of a word with each of the eleven basic colours in WordNet
can be used to automatically determine the colour most associated with the 177 thesaurus categories
from the gold standard described in Section 5 earlier.
% having strong color associations.
% for which we have gold standard associations with colours.
We determine closeness using two similarity measures---Jiang and Conrath \shortcite{JiangC97} and Lin \shortcite{Lin97}---and two relatedness measures---Lesk \cite{BanerjeeP03} and gloss vector overlap \cite{PedersenPM04a}---from the
WordNet Similarity package.

For each thesaurus category--colour pair, we summed the WordNet closeness of each of the terms in the category to
the colour. The colour with the highest sum is chosen as the one closest to the thesaurus category.
Section (c) and section (d) of Table~\ref{tab:auto results}, show how often the closest colours are also the colours most associated with the gold standard categories.
Section (a) lists some unsupervised baselines. Random-choice baseline is the score obtained when a colour is chosen at random (1/11 = 9.1\%).
Another baseline is a system that always chooses the most frequent colour in a corpus.
Section (a) reports three such baseline scores obtained by choosing the most frequently occurring colour in three separate corpora.
Section (b) lists a supervised baseline obtained by choosing the colour most commonly associated with a categories in the gold standard.
The automatic measures listed in sections (c) through (f) do not have access to this information.

Observe that the relatedness measures  %of gloss vector and Lesk 
are markedly better
than the similarity measures
at identifying the true associated colour.
Yet, for a majority of the thesaurus categories 
the closest colour in WordNet is not the most associated colour.
% This is not surprising since WordNet was not created with the specific intention of comprehensively linking concepts with associated colours. 
% We hope that our crowdsourced word--colour lexicon will add value to WordNet in this regard.

\subsection{Co-occurrence in Text}
Physical entities that tend to have a certain colour tend to be associated with that colour. For example leaves are associated with green.
Intuition suggests that these entities will co-occur with the associated colours more often than with any other colour.
As language has expressions such as {\it green with envy} and {\it feeling blue}, we also expect
that certain abstract notions, such as {\it envy} and {\it sadness}, will co-occur more often with their
associated colours, green and blue respectively, more often than with any other colour.
We now describe experiments to determine the extent to which target concepts
co-occur in text most often with their associated colours.

We selected three corpora to investigate occurrences of colour terms: the {\it British National Corpus (BNC)} \cite{Burnard00u}, 
the {\it Google N-gram Corpus (GNC)}, and the {\it Google Books Corpus (GBC)} \cite{GBC}.\footnote{The {\it BNC}
is available at: http://www.natcorp.ox.ac.uk.\\ The {\it GNC} is available through the Linguistic Data Consortium.\\ The {\it GBC} is available at http://ngrams.googlelabs.com/datasets.}
The {\it BNC}, a 100 million word corpus, is considered to be fairly balanced with text from various domains.
% It has a 100million words, and so by today's standards perhaps a little small.
The {\it GNC} is a trillion-word web coprus.
The {\it GBC} is a digitized version of about 5.2 million books, % (close to 4\% of all published books).
and the English portion has about 361 billion words.
The {\it GNC} and {\it GBC} are distributed as collections of 1-gram to 5-gram files. %unigram, bigram, trigram, fourgram, and fivegram frequencies.

Table \ref{tab:uni colours} shows the frequencies and ranks of the eleven basic colour terms in the {\it BNC} and the unigram files
of {\it GNC} and {\it GBC}. The ranking is from the most frequent to the least frequent colour in the corpus.
The last column lists the Spearman rank order correlation ($\rho$) of the rankings with the Berlin and Kay ranks \shortcite{BerlinK69} (listed in Section 2:(1)).
Observe that order of the colours from most frequent to least frequent in the {\it GNC} and {\it GBC} have a strong correlation
with the order proposed by Berlin and Kay, especially so for the rankings obtained from counts in the {\it Google Books Corpus}.

\begin{table}[t]
% \caption{colour agreement.}

\centering
 \resizebox{0.5\textwidth}{!}{
% {\small
\begin{tabular}{lr}
\hline
                    Automatic method for choosing colour     &Accuracy   \\
\hline
		(a) Unsupervised baselines: \rule{0pt}{10pt} & \\
		$\;\;\;\;\;$ - randomly choosing a colour 			&9.1\\
		$\;\;\;\;\;$ - most frequent colour in {\it BNC} (black)			&23.2\\
		$\;\;\;\;\;$ - most frequent colour in {\it GNC} (black)			&23.2\\
		$\;\;\;\;\;$ - most frequent colour in {\it GBC} (white)		&33.3\\
		(b) Supervised baseline: \rule{0pt}{10pt} & \\
		\multicolumn{2}{l}{$\;\;\;\;\;$ - colour most often associated }\\
		$\;\;\;\;\;\;\;$ with categories (white) 									&33.3\\
        (c) WordNet similarity measures: \rule{0pt}{10pt} & \\ %\rule{0pt}{14pt} 					&       &\\
%		$\;\;\;\;\;\;\;\;$ - {Similarity measures:}					& 		&\\
		$\;\;\;\;\;$ - Jiang Conrath measure 			 	&15.7\\
		$\;\;\;\;\;$ - Lin's measure 					 	&15.7\\
        (d) WordNet relatedness measures: \rule{0pt}{10pt} & \\ %\rule{0pt}{14pt} 					&       &\\
		% $\;\;\;\;\;\;\;\;\;\;\;\;$ path measure 					&9.1 	&6.5\\
%		$\;\;\;\;\;\;\;\;$ - {Relatedness measures:}				& 		&\\
		$\;\;\;\;\;$ - Lesk measure 					 	&24.7\\
		$\;\;\;\;\;$ - gloss vector measure 			 	&28.6\\
        (e) Co-occurrence in text: \rule{0pt}{10pt}        							&       \\
%		$\;\;\;\;\;\;\;\;$ - {Google N-gram Corpus:}				& 		&\\
%		$\;\;\;\;\;\;\;\;\;\;\;\;$ - conditional probability 		&19.2 	&26.8\\
		$\;\;\;\;\;$ - $p(colour|word)$ in {\it BNC}  				&31.4\\
		$\;\;\;\;\;$ - $p(colour|word)$ in {\it GNC}  			&37.9\\
%		$\;\;\;\;\;\;\;\;$ - {Google Books Corpus}					& 		&\\
%		$\;\;\;\;\;\;\;\;\;\;\;\;$ - conditional probability 		&25.4 	&32.5\\
		$\;\;\;\;\;$ - $p(colour|word)$ in {\it GBC}  				&38.3\\
        \multicolumn{2}{l}{(f) Co-occurrence and polarity: \rule{0pt}{10pt}}\\
%		$\;\;\;\;\;\;\;\;$ - {Google N-gram Corpus:}				& 		&\\
%		$\;\;\;\;\;\;\;\;\;\;\;\;$ - conditional probability 		&26.2 	&37.2\\
		$\;\;\;\;\;$ - $p(colour|word,polarity)$ in {\it BNC}  							&51.4\\
		$\;\;\;\;\;$ - $p(colour|word,polarity)$ in {\it GNC}  				&47.6\\
%		$\;\;\;\;\;\;\;\;$ - {Google Books Corpus}					& 		&\\
%		$\;\;\;\;\;\;\;\;\;\;\;\;$ - conditional probability 		&32.7	&45.6\\
		$\;\;\;\;\;$ - $p(colour|word,polarity)$ in {\it GBC}  				&{\bf 60.1}\\
 
\hline
\end{tabular}
 }
\label{tab:auto results}
\vspace*{-3mm}
\caption{Percentage of times the colour chosen by automatic method is also the
colour identified by annotators as most associated to a thesaurus category.
% {\it BNC} is the {\it British National Corpus},
% {\it GNC} is the {\it Google N-gram Corpus}, and {\it GBC} is the {\it Google Books Corpus}.
}
\vspace*{-2mm}
\end{table}

For each of the 177 gold standard thesaurus categories, % having strong colour association (as calculated in Section 5),
we determined the conditional probability of co-occurring with different colour terms in the {\it BNC, GNC,} and {\it GBC}. 
The total co-occurrence frequency of a category with a colour was calculated by summing up the co-occurrence
frequency of each of the terms in it with the colour term. 
We used a four-word window as context.  The counts from {\it GNC} and {\it GBC} were determined using the fivegram files.
% in the fivegram file, and used it
%to calculate the conditional probability of the the colour given the category.
Section (e) in Table~\ref{tab:auto results} shows how often the colour with the highest conditional probability is also the colour most associated with a category.
These numbers are higher than the baselines (a and b), as well as the scores obtained by the WordNet approaches (c).
% but they are not high enough to be consistently good predictors of a category's associated colour.

From Table 5 in  Section 7, we know that some colours tend to be strongly positive and others negative.
We wanted to determine how useful these polarity cues can be in identifying 
the colour most associated with a category. We used the automatically generated Macquarie Semantic Orientation Lexicon (MSOL) \cite{MohammadDD09}
to determine if a thesaurus category is positive or negative.\footnote{MSOL is available at http://www.umiacs.umd.edu/$\sim$saif/\\WebPages/ResearchInterests.html\#semanticorientation.}
A category is marked as negative if it has more negative words than positive, otherwise it is marked as positive.
If a category is positive, then co-occurrence cues were used to select a colour from only the positive colours (white, green, yellow,
blue, pink, and orange), whereas if a category is negative, then co-occurrence cues select from only the negative colours (black, red, brown,
and grey). %\footnote{More sophisticated ways to combine polarity with co-occurrence cues may be employed, but we were interested
% here in determining some minimum improvement the polarity cues can provide.}
Section (f) of Table~\ref{tab:auto results} provides results with this method.
Observe that these numbers are a marked improvement over Section (e) numbers, suggesting
that polarity cues can be very useful in determining concept--colour association.

Counts from the {\it GNC} yielded poorer results compared to the much smaller {\it BNC}, and 
the somewhat smaller {\it GBC} possibly because frequency counts from {\it GNC} are available
only for those n-grams that occur at least thirty times. 
% The long tail of low frequency co-occurrences may have helped the BNC- and GBC-based systems. 
Further, {\it GBC} and {\it BNC} are both collections of edited texts, and so 
expected to be cleaner than the {\it GNC} which is a corpus extracted from the World Wide Web.

% \begin{figure*}[]
%  \begin{center}
%  \includegraphics[width=1.35\columnwidth]{band.png}
%  \end{center}
%  % \caption{{\bf Polarity pie chart}---Proportions of polarities in quotations about gratitude. (Text from: Quote Garden)}
%  % \caption{Scatter plot of imageability and association with colour of thesaurus categories.}
%   % \vspace*{-4mm}
%  \caption{Scatter plot of thesaurus categories. The area of high colour association is shaded.}
%  \label{fig:scatter-auto}
%  % \vspace*{-4mm}
%  \end{figure*}

% \subsection{Discussion}
% \begin{itemize}
% \item The higher than random choice results of row (b) and row (c) in Table~\ref{tab:auto results} suggest that both closeness in WordNet and co-occurrence in text 
% are cues to association between a concept and a colour. However, these scores are not high-enough to be good predictors
% of the colour most associated with a concept. 
% \item More sophisticated approaches than simple co-occurrence must be developed to detect word-colour association from text and semantic resources, such as WordNet.

% \end{itemize}

\section{Conclusions and Future Work}
We created a large word--colour association lexicon by crowdsourcing, which we will make publicly available.
Word-choice questions were used to guide the annotators to the desired senses of the target words, and also as a gold
questions for identifying malicious annotators (a common problem in
crowdsourcing). % detect and discard erroneous input.
We found that more than 32\% of the words and 33\% of the {\it Macquarie Thesaurus} categories have a strong association with one of the eleven colours
 chosen for the experiment. 
% These numbers show that for a substantial number of concepts there is consensus for associated colour.
We analyzed abstract concepts, emotions in particular, and showed that they
too have strong colour associations. % , especially emotions.
Thus, using the right colours in tasks such as information visualization and web development, can not only improve semantic coherence but also
inspire the desired emotional response.

Interestingly, we found that frequencies of colour associations follow the same order in which colour terms occur in different languages \cite{BerlinK69}.
The frequency-based ranking of colour terms in the 
{\it BNC, GNC}, and {\it GBC} also had a high correlation with the Berlin and Kay order.
% We also found the {\it BNC} to have a significantly higher number of colour terms for every million words, than the other two corpora.

Finally, we show that automatic methods that rely on co-occurrence and polarity cues alone, and
no labeled information of word--colour association,
can accurately estimate the colour associated with a concept more than 60\% of the time. The random choice and supervised
baselines for this task are 9.1\% and 33.3\%, respectively.
% We expect semi-supervised approaches that use seed word--colour associations to obtain even better results.
% Such approaches can be used to further improve the coverage of the manually created lexicon, and for creating
% similar lexicons in other languages.
% We showed that co-occurrence cues are useful in automatically detecting the colour most associated with a thesaurus category,
% but they markedly more accurate when used in tandem with information about the polarity of the category.
% Counts from the recently-released {\it Google Books Corpus} gave us the best results, and we believe this is because of its large size (much bigger than {\it BNC}),
% better linguistic quality (edited text from books, as opposed to extracts from the World Wide Web), and availability of the
% complete list of fivegram frequencies (unlike the {\it GNC} that has entries for only those fivegrams that occur thirty times or more).
% We are now in the process of doubling the size of the word--colour lexicon through even more crowdsourced annotations.
% We also intend to identify cross-cultural and cross-lingual differences in word--colour association.
% Automatic methods to determine word--colour association can be further improved by correlating colours in images
% with terms listed in their captions.
We are interested in using word--colour associations as a feature in sentiment analysis and for paraphrasing.

%  \section*{Acknowledgments}
% We thank the reviewers in advance for their time and thoughtful comments.
% 

% \vspace*{-1mm}
\section*{Acknowledgments}
\vspace*{-2mm}
This research was funded by the National Research Council Canada.
Grateful thanks to Peter Turney, Tara Small, and the reviewers for many wonderful ideas.
 Thanks to the thousands of people who answered the colour survey with diligence and care.

\bibliography{references}

\end{document}